\pdfoutput=1

\documentclass[11pt]{article}

\usepackage{emnlp2021}

\usepackage{times}
\usepackage{latexsym}
\usepackage{graphicx}
\usepackage{amsmath}
\usepackage{amssymb}
\usepackage{booktabs}
\usepackage{multirow}
\usepackage{array}
\usepackage{enumitem}
\usepackage{subfigure}
\usepackage[T1]{fontenc}
\allowdisplaybreaks

\usepackage[utf8]{inputenc}

\usepackage{microtype}

\usepackage{hyperref}

\title{Rethinking Zero-shot Neural Machine Translation: \\From a Perspective of Latent Variables}

\author{
    Weizhi Wang$^{1}$\thanks{\ \ Contribution during internship at Alibaba.}, Zhirui Zhang$^{2}$\thanks{\ \ Corresponding author.}, Yichao Du$^{3}$, Boxing Chen$^{2}$, Jun Xie$^{2}$, and Weihua Luo$^{2}$ \\
    $^{1}$Rutgers University, New Brunswick, USA\\
    $^{2}$Machine Intelligence Technology Lab, Alibaba DAMO Academy \\ 
    $^{3}$University of Science and Technology of China, China \\
    \texttt{$^{1}$weizhi.wang@rutgers.edu} \ \texttt{$^{2}$zrustc11@gmail.com} \\
    \texttt{$^{2}$\{boxing.cbx, qingjing.xj, weihua.luowh\}@alibaba-inc.com} \\ 
    \texttt{$^{3}$duyichao@mail.ustc.edu.cn}
}

\begin{document}
\maketitle
\begin{abstract}
Zero-shot translation, directly translating between language pairs unseen in training, is a promising capability of multilingual neural machine translation (NMT).
However, it usually suffers from capturing spurious correlations between the output language and language invariant semantics due to the maximum likelihood training objective, leading to poor transfer performance on zero-shot translation.
In this paper, we introduce a denoising autoencoder objective based on pivot language into traditional training objective to improve the translation accuracy on zero-shot directions.
The theoretical analysis from the perspective of latent variables shows that our approach actually implicitly maximizes the probability distributions for zero-shot directions.
On two benchmark machine translation datasets, we demonstrate that the proposed method is able to effectively eliminate the spurious correlations and significantly outperforms state-of-the-art methods with a remarkable performance. Our code is available at \url{https://github.com/Victorwz/zs-nmt-dae}.

\end{abstract}

\section{Introduction}
Multilingual neural machine translation (NMT) system concatenates multiple language pairs into one single neural-based model, enabling translation on multiple language directions~\cite{Firat2016MultiWayMN,Ha2016TowardMN,Johnson2017GooglesMN,Kudugunta2019InvestigatingMN,Arivazhagan2019MassivelyMN,Zhang2020ImprovingMM}. 
Besides, the multilingual NMT system can achieve translation on unseen language pairs in training, and we refer to this setting as zero-shot NMT. 
This finding is promising that zero-shot translation halves the decoding time of pivot-based method and avoids the problem of error propagation.
Meanwhile, zero-shot NMT casts off the requirement of parallel data for a potentially quadratic number of language pairs, which is sometimes impractical especially between low-resource languages.
Despite the potential benefits, achieving high-quality zero-shot translation is a very challenging task.
Standard multilingual NMT systems are sensitive to hyper-parameter settings and tend to generate poor outputs.

One line of research believes that the success of zero-shot translation depends on the ability of the model to learn language invariant features, or an interlingua, for cross-lingual transfer \cite{Arivazhagan2019TheMI,Ji2020CrosslingualPB,Liu2021ImprovingZT}.
\citet{Arivazhagan2019TheMI} design auxiliary losses on the NMT encoder
that impose representational invariance across languages. \citet{Ji2020CrosslingualPB} build up a universal encoder for different languages via bridge language model pre-training, while \citet{Liu2021ImprovingZT} disentangle positional information in multilingual NMT to obtain language-agnostic representations.
Besides, \citet{Gu2019ImprovedZN} point out that the conventional multilingual NMT model heavily captures spurious correlations between the output language and language invariant semantics due to the maximum likelihood training objective, making it hard to generate a reasonable translation in an unseen language.
Then they investigate the effectiveness of decoder pre-training and back-translation on this problem. 

\begin{table}[tb]\small
\centering
\begin{tabular}{@{}c|c}
\toprule
Model     & BLEU on DE$\Rightarrow$FR    \\ \midrule
DE$\Rightarrow$EN+EN$\Rightarrow$FR   & $6.0$  \\ \midrule
PIV-(DE$\Rightarrow$EN+EN$\Rightarrow$FR) & $31.7$ \\ \bottomrule
\end{tabular}
\vspace{-5pt}
\caption{BLEU scores [\%] of training multilingual NMT with these two translation directions and its pivoting variant on Europarl Dataset.
}
\label{table:problem}
\end{table}

In this paper, we focus on English-centric multilingual NMT and propose to incorporate a simple denoising autoencoder objective based on English language into the traditional training objective of multilingual NMT to achieve better performance on zero-shot directions.
This approach is motivated by an observation that:
as shown in Table \ref{table:problem}, if we only optimize two translation directions DE$\Rightarrow$EN and EN$\Rightarrow$FR in a single model, it hardly achieves successful zero-shot translation on DE$\Rightarrow$FR.
It is because that the model easily learns high mutual information between language semantics of German and output language, ignoring the functionality of language IDs.
Actually, this mutual information can be significantly alleviated by directly replacing the original German sentence with a noisy target English sentence in training data, thereby guiding the model to learn the correct mapping between language IDs and output language.
Besides, we analyze our proposed method by treating pivot language as latent variables and find that our approach actually implicitly maximizes the probability distributions for zero-shot translation directions.

We evaluate the proposed method on two public multilingual datasets with several English-centric language-pairs, Europarl \cite{koehn2005europarl} and MultiUN \cite{Ziemski2016TheUN}.
Experimental results demonstrate that our proposed method not only achieves significant improvement over vanilla multilingual NMT on zero-shot directions, but also outperforms previous state-of-the-art methods.

\section{Multilingual NMT}
The multilingual NMT system~\cite{Johnson2017GooglesMN} combines different language directions into one single translation model.
Due to data limitations of non-English languages, multilingual NMT systems are mostly trained on large-scale English-centric corpus via maximizing the likelihood over all available language pairs $\mathcal{S}$:
\begin{equation}
    \mathcal{L}_{m}(\theta) = \sum_{(i,j)\in \mathcal{S},(x,y)\in D^{i,j}}\log P(y|x,\mathbf{j};\theta),
\end{equation}
where $(i,j)\in \mathcal{S}$ are the sampled source language ID and target language ID in all available language pairs, $D^{i,j}$ represents for the corresponding parallel data, and $\theta$ is the model parameter. 
The target language ID is appended as the initial token of source sentences, to let the model know which language it should translate to. 
In addition, the multilingual NMT system has proven the capability of translating on unseen pairs in training~\cite{Firat2016MultiWayMN, Johnson2017GooglesMN}, which is a property of \textbf{zero-shot translation}. 
However, the zero-shot translation quality significantly falls behind that of pivoting methods. 
The main issue leading to the unsatisfactory performance is that the multilingual NMT model captures spurious correlations between the output language and language invariant semantics due to the maximum likelihood training objective~\cite{Gu2019ImprovedZN}.

\section{Method}
In this section, we first introduce the denoising autoencoder task and then analyze the effectiveness of our proposed method from the perspective of latent variables.

\paragraph{Denoising Autoencoder Task.}
Given English-centric parallel data (X/Y/...$\Leftrightarrow$EN), we usually optimize the maximum likelihood training objective to build the multilingual NMT model.
Since the target language ID is inserted at the beginning of the source sentence and only treated as a single token, the maximum likelihood training objective easily ignores the functionality of target language ID, leading to unreasonable mutual information between language semantic of ``X/Y/...'' and output language of English.
To address this problem, we introduce a denoising sequence-to-sequence task, in which we directly replace the original input sentence with a noisy target English sentence in training data. 
In this way, previous mutual information can be significantly reduced, while enhancing the relationship between language IDs and output language.
Specifically, we simply use all English sentences in parallel data to construct the denoising English corpus $D_{\text{EN}}$ via text infilling operation~\cite{Lewis2020BARTDS}. 
Then we optimize the multilingual NMT model via maximizing the original translation objective $\mathcal{L}_{m}(\theta)$ and denoising autoencoder objective $\mathcal{L}_{d}(\theta)$:
\begin{align}
 \mathcal{L}_{d}(\theta) &= \sum_{j=\text{<2en>}, (\overline{y}, y)\in D_{\text{EN}}}\log P(y|\overline{y},\mathbf{j};\theta), \\
\mathcal{L}_{a}(\theta) &=\mathcal{L}_{m}(\theta) + \mathcal{L}_{d}(\theta).
\end{align}

\paragraph{Latent Variable Perspective.}
As for zero-shot translation, we actually aim at directly fitting the probability distribution between non-English languages ``X/Y/...'' in the unified multilingual NMT system.
For convenience, we consider the probability distribution $P(Y|X;D^{*})$ between two non-English languages over the ideal parallel training data $D^{*}$.
In practice, it is difficult to obtain such training data $D^{*}$ for the model training.
To handle this issue, we convert the task of maximizing $P(Y|X;D^{*})$ into optimizing three existing sub-tasks, by treating the English language as a latent variable $h$ and introducing the probability distribution $P(h|\overline{h})$ of denoising autoencoder task:
\begin{equation}
\begin{split}
    &P(Y|X;D^{*}) = \sum_{(x,y) \in D^{*}}{\log P(y|x)} \\
    & = \sum_{(x,y) \in D^{*}}{\log \sum_{h}{P(y|h,x)P(h|x)}} \\
    & \approx \sum_{(x,y) \in D^{*}}{\log \sum_{h} P(h|\overline{h}) \frac{P(y|h)P(h|x)}{P(h|\overline{h})} }  \\
    & \geq \sum_{(x,y) \in D^{*}}{\sum_{h} P(h|\overline{h}) \log \frac{P(y|h)P(h|x)}{P(h|\overline{h})} }  \\
    & = \sum_{(x,y) \in D^{*}} {\mathbb{E}_{h \sim P(h|\overline{h})}\log P(y|h)}   \\
    & \ \ \ \ \ \ \ \ \ \ \ \ \ \ \ \ \ \ \ \ \ \ \ \ \ \ \ - \mathbb{KL}(P(h|\overline{h})||P(h|x))   \\
    & = P^{*}(Y|X;D^{*},P(h|\overline{h})), 
\label{equ:elbo}
\end{split}
\end{equation}
where we assume that $P(y|h,x)\approx P(y|h)$ due to the semantic equivalence of languages $h$ and $x$.
With above equation, the original objective is transformed into optimizing three sub-tasks $P(h|x)$, $P(y|h)$ and $P(h|\overline{h})$. 
Incorporating the denoising autoencoder objective into the translation objective of multilingual NMT model helps minimize the KL-divergence terms, thus implicitly maximizing the lower bound of probability distributions of zero-shot directions.
Following \citet{Ren2018TriangularAF}, the gap between $P^{*}(Y|X;D^{*},P(h|\overline{h}))$ and $P(Y|X;D^{*})$ can be calculated as follow: 
\begin{align}
&\Delta:= P(Y|X;D^{*}) - P^{*}(Y|X;D^{*},P(h|\overline{h}))  \nonumber \\
& = \sum_{(x,y) \in D^{*}}{\sum_{h} P(h|\overline{h}) \log \frac{P(h|\overline{h}) P(y|x)}{P(y|h) P(h|x)}} \nonumber \\
& \approx \sum_{(x,y) \in D^{*}} \mathbb{KL}(P(h|\overline{h})||P(h|y)), 
\label{equ:delta}
\end{align}
where we leverage an additional approximation that $P(h|x,y)\approx P(h|y)$ due to the semantic equivalence.
Refer to Appendix \ref{appendix_2} for detailed derivations.
Once we complement $P(h|y)$ into three sub-tasks mentioned before, this gap could be further reduced, resulting in better performance on zero-shot translation directions.

\section{Experiments}

\subsection{Experimental Settings}
\paragraph{Datasets.} 
We evaluate the proposed method on two benchmark machine translation datasets, Europarl and MultiUN. The data statistics of two selected datasets are summarized in Table~\ref{tab:datasize}.
BLEU \cite{Papineni2002BleuAM} is used as the metric for evaluating translation quality.
For Europarl dataset, we select three European languages, Germany (De), French (Fr) and English (En).
We remove all parallel sentences between De and Fr to ensure the zero-shot setting.
We use WMT \textit{devtest2006} as validation set and \textit{test2006} as test set. 
For MultiUN, four languages are selected, Arabic (Ar), Chinese (Zh), Russian (Ru), and English (En).
The selected languages are distributed in various language families, making the zero-shot language transfer more difficult. 
We use MultiUN standard validation and test sets to report the zero-shot performance. 
To differentiate language pairs, we follow \citet{Johnson2017GooglesMN} to append the language tag ``<2Y>'' on the source side for translating $X\Rightarrow Y$. 

\begin{table}[t]\small
\centering
\begin{tabular}{l|ccc}
\toprule
Dataset  & Language Pairs & Train & Dev \& Test  \\
\midrule
Europarl & De-En, Fr-En & 1.8M & 2000 \\
\midrule
MultiUN & Ar-En, Zh-En, Ru-En & 2M & 4000 \\
\bottomrule
\end{tabular}
\vspace{-5pt}
\caption{\label{tab:datasize}Data statistics of Europarl and MultiUN, in which we sub-sampled 2M samples for each language-pair in MultiUN. }
\end{table}

\begin{table*}[ht]\small
\centering
\begin{tabular}{lp{1cm}<{\centering}p{1cm}<{\centering}p{1cm}<{\centering}p{1cm}<{\centering}p{1cm}<{\centering}p{1cm}<{\centering}cc}
\toprule
MultiUN & \multicolumn{8}{c}{Ar, Zh, Ru $\leftrightarrow$ En} \\
\midrule
\multirow{2}*{Model} & \multicolumn{2}{c}{Ar-Ru} & \multicolumn{2}{c}{Ar-Zh} &  \multicolumn{2}{c}{Ru-Zh} & Zero & Parallel \\
~ & $\leftarrow$ & $\rightarrow$ & $\leftarrow$ & $\rightarrow$ & $\leftarrow$ & $\rightarrow$ & Avg. & Avg. \\
\midrule
MNMT & 17.9 & 	13.4 & 	16.1 & 	29.5 &	12.1 &	30.3 &	19.9 &	49.2 \\
LM+MNMT & 22.0 &  29.3 &	20.3 &	42.7 &	24.3 &	42.1 &	30.1 &	48.9 \\
MNMT-RS & 20.8 & 26.1 & 20.3 & 37.9 & 24.2 & 37.4 & 27.8 & 49.9 \\
\midrule
MNMT+DN (Ours) & \textbf{24.6} & \textbf{33.0} & \textbf{24.6} & \textbf{47.2} & \textbf{30.0} & \textbf{46.1} &	\textbf{34.3} & \textbf{50.1} \\
\bottomrule
\end{tabular}
\vspace{-5pt}
\caption{\label{tab:multiun}Overall BLEU scores [\%] on six zero-shot directions of MultiUN dataset. ``Zero Avg.'' and ``Parallel Avg.'' refer to average BLEU score of six zero-shot directions and six supervised directions, respectively.}
\end{table*}

\paragraph{Baselines.} 
In our experiments, we compare the proposed method \textbf{MNMT+DN} with the following approaches: (\textit{i}) \textbf{MNMT} \cite{Johnson2017GooglesMN}: training a multilingual NMT model on all directions with available parallel data; (\textit{ii}) \textbf{LM+MNMT} \cite{Gu2019ImprovedZN}: pre-training the decoder as a multilingual language model, then training the MNMT model initialized with the pre-trained decoder; (\textit{iii}) \textbf{MNMT-RC} \cite{Liu2021ImprovingZT}: removing residual connections in an encoder layer to disentangle positional information.
We re-implement all baseline methods, following the same experimental settings to make fair comparison with our method.

\begin{figure}[t] 
\centering 
\includegraphics[width=0.46\textwidth]{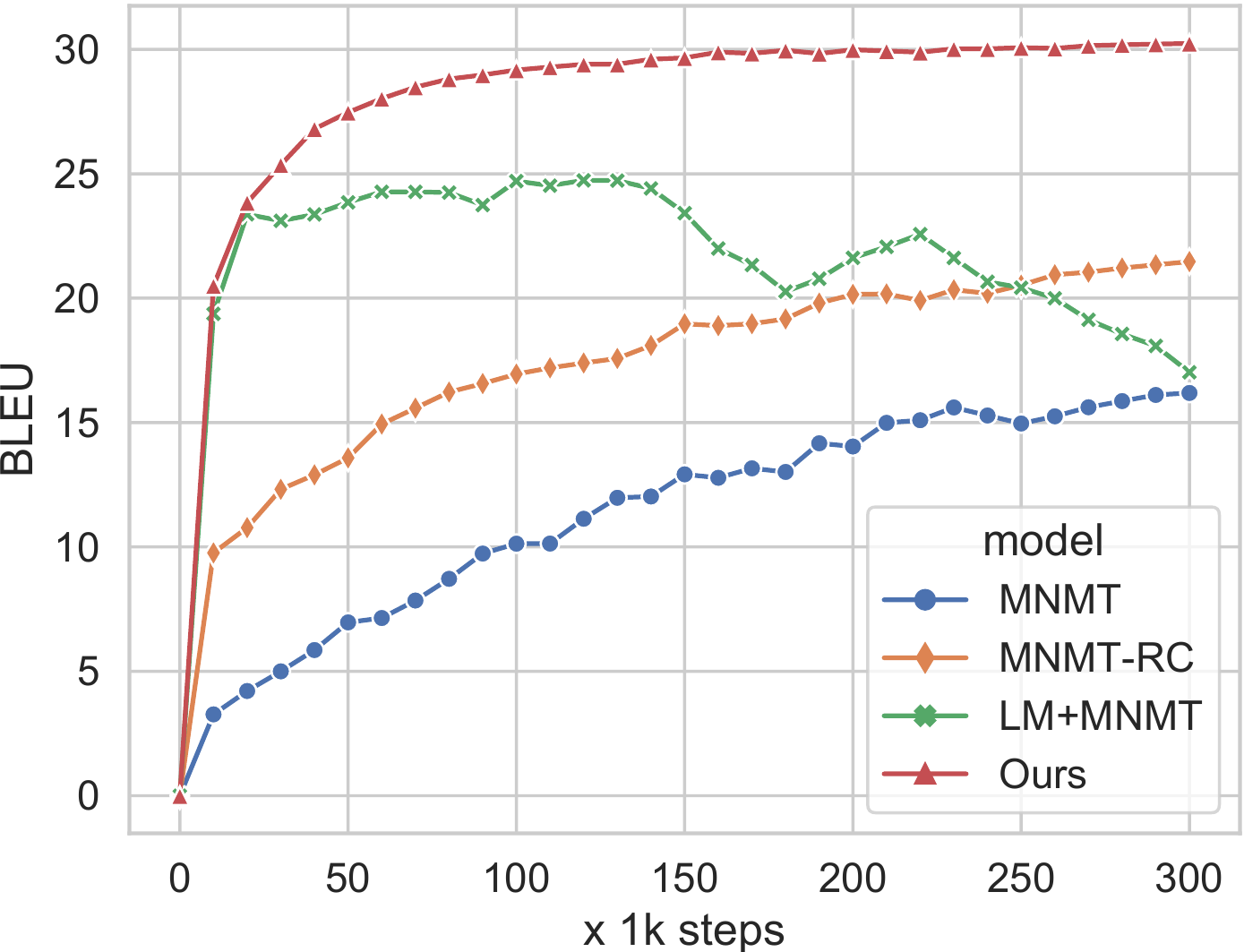}
\vspace{-5pt}
\caption{Learning curve of different methods on MultiUN dataset. We sub-sample 1K sentences from every zero-shot translation direction and report BLEU score on the combined 6K-size validation set. }
\label{fig:curve}
\end{figure}

\paragraph{Experimental Details.}
\label{sec:details}
We choose standard Transformer-base~\cite{Vaswani2017AttentionIA} architecture to conduct experiments on all baseline and proposed methods, with $n_{\text{layer}}=6,n_{\text{head}}=8, d_{\text{embed}}=512$. 
We use \texttt{faiseq} toolkit\footnote{\url{https://github.com/pytorch/fairseq}} \cite{Ott2019fairseqAF} for fast implementations and experiments. 
We deploy Adam \cite{Kingma2015AdamAM} ($\beta_1=0.9,\beta_2=0.98$) optimizer and train all models with $lr=0.0005, t_{\textnormal{warmup}}=4000, \textnormal{dropout}=0.1, n_{\textnormal{batch}}=8000\ \textnormal{tokens}$. 
The \texttt{Moses} toolkit\footnote{\url{https://github.com/moses-smt/mosesdecoder}} \cite{koehn2007moses} is used to tokenize translation corpus. Exceptionally, we use Jieba\footnote{\url{https://github.com/fxsjy/jieba}} for Chinese tokenization.
For each dataset, we lowercase all data and preprocess the corpus with 40K Byte-Pair-Encoding (BPE) \cite{Sennrich2016NeuralMT} operations on all languages.
For our proposed approach, we mask 30\% of tokens in the whole training corpus, and deploy span masking \cite{Joshi2020SpanBERTIP}, in which a sequence text spans are sampled and masked, with the masked span lengths sampled from a Poisson distribution ($\lambda$ = 3). 
0-length spans correspond to the insertion of \texttt{[MASK]} token.
Every model is trained for 300k updates on Europarl or 500K updates on MultiUN (additional 100k updates for pre-training), and the best model is selected based on BLEU score on validation set every 10k updates. For decoding, we adopt beam-search with beam size = 5 and calculate BLEU scores using \texttt{SacreBLEU}\footnote{\url{https://github.com/mjpost/sacrebleu}}.

\subsection{Results on MultiUN Dataset}

Table~\ref{tab:multiun} reports the main results on the MultiUN dataset. 
We can find that our proposed method achieves state-of-the-art performance on all six zero-shot translation directions among all multilingual NMT systems.
In addition, our method significantly improves the zero-shot performance of vanilla MNMT model by an average 14.4 BLEU score without performance degradation on supervised directions. 
These results demonstrate the effectiveness of incorporating denoising autoencoder objective in the training of multilingual NMT.
We further investigate the learning curve of different methods on the validation set.
As shown in Figure~\ref{fig:curve}, our proposed method reaches faster convergence than MNMT and MNMT-RC, while LM+MNMT easily leads to over-fitting.

\subsection{Results on Europarl Dataset}
The main results on the Europarl dataset are presented in Table~\ref{tab:europarl}.
We can observe that our proposed method still significantly improves the zero-shot translation performance of multilingual NMT systems with an average of 5.1 BLEU score improvements.
Different from the MultiUN dataset with four languages distributed in different language families, the selected languages (De, Fr, En) of Europarl are all European languages, making the gap between various baselines and our method smaller than that of MultiUN.

\begin{table}[t]\small
\centering
\begin{tabular}{lp{0.6cm}<{\centering}p{0.6cm}<{\centering}cc}
\toprule
Europarl & \multicolumn{4}{c}{De, Fr $\leftrightarrow$ En} \\
\midrule
\multirow{2}*{Model} & \multicolumn{2}{c}{De-Fr}  & Zero & Parallel \\
~ & $\leftarrow$ & $\rightarrow$  & Avg. & Avg. \\
\midrule
MNMT & 21.5 & 27.3 & 24.4 & 34.1 \\
LM+MNMT & 25.5 & 31.1 &  28.3 & 33.6 \\
MNMT-RC & 25.1 & 30.8 &  28.0 & 33.5 \\
\midrule
MNMT+DN (Ours) & \textbf{27.1} & \textbf{31.8} &  \textbf{29.5} & \textbf{33.7} \\
\bottomrule
\end{tabular}
\vspace{-5pt}
\caption{\label{tab:europarl}Overall BLEU scores [\%] on two zero-shot directions of Europarl dataset.} 
\end{table}

\subsection{Evaluation of Off-Target Translations}
We further summarize the percentage of off-target translations on zero-shot directions to verify the effectiveness of the proposed method. 
Generating off-target translations means that the multilingual NMT system fails in achieving zero-shot translation and generates translation in wrong output language. 
We use \texttt{langdetect}\footnote{\url{https://github.com/Mimino666/langdetect}} toolkit to capture the off-target translations and calculate the language accuracy as $(1-n_\text{off-target}/n_\text{sentences})$. The results of language accuracy on two selected corpora are presented in Table~\ref{tab:accuracy}. The proposed method achieves the language accuracy of 99.13\% on Europarl and 95.76\% on MultiUN, which surpass baseline methods with a significant improvement. 
The results demonstrate that our method effectively alleviates the issue of off-target translation in zero-shot directions.

\begin{table}[t]\small
\centering
\begin{tabular}{lcc}
\toprule
\multirow{2}*{Model} & \multicolumn{2}{c}{Dataset} \\
~ & MultiUN & Europarl \\
\midrule
MNMT & 57.49\% & 98.19\%   \\
LM+MNMT & 91.87\% & 99.13\%   \\
MNMT-RC & 83.37\% & 99.00\%  \\
\midrule
MNMT+DN (Ours) & \textbf{95.76\%} & \textbf{99.13\%}  \\
\bottomrule
\end{tabular}
\vspace{-5pt}
\caption{\label{tab:accuracy}Average language accuracy on all zero-shot directions of two selected datasets.}
\end{table}

\subsection{Ablation Study}
As illustrated in Equation~\ref{equ:elbo}, the training objective of zero-shot directions can be converted into optimizing three sub-tasks jointly. 
To verify this analysis, we conduct an ablation study on the Europarl dataset.
We consider a single model with two translation directions DE$\Rightarrow$EN+EN$\Rightarrow$FR.
As shown in Table~\ref{tab:ablation}, when incorporating denoising autoencoder task, DE$\Rightarrow$EN+EN$\Rightarrow$FR+DN achieves a remarkable zero-shot performance on DE$\Rightarrow$FR of 31.1 BLEU score.
This result demonstrates that the introduction of denoising autoencoder task can effectively break the spurious correlations between output language and semantics, enabling the failed model to perform zero-shot translation. 
Complementing with more translation tasks, such as FR$\Rightarrow$EN and EN$\Rightarrow$DE, MNMT+DN further improves translation accuracy on DE$\Rightarrow$FR, which proves the analysis of Equation~\ref{equ:delta}.
In addition, an alternative to our proposed method is BART pre-training (BART-PT), which first learns the denoising autoencoder objective and fine-tunes on the multilingual corpus. 
We can observe that BART-PT gains a similar performance to LM+MNMT, but worse than MNMT+DN due to the catastrophic forgetting problem~\cite{McCloskey1989CatastrophicII}.
The full results of BART-PT on MultiUN and Europarl datasets are illustrated in Appendix \ref{full_result}.

\begin{table}[t]\small
\centering
\begin{tabular}{lcccc}
\toprule
Europarl & \multicolumn{4}{c}{De, Fr $\leftrightarrow$ En} \\
\midrule
\multirow{2}*{Setting} & \multicolumn{2}{c}{De-Fr}  & Zero & Parallel  \\
~ & $\leftarrow$ & $\rightarrow$  & Avg. & Avg.\\
\midrule
DE$\Rightarrow$EN+EN$\Rightarrow$FR  & - & 6.0 & - & - \\
DE$\Rightarrow$EN+EN$\Rightarrow$FR+DN & - & 31.1 & - & - \\
MNMT & 21.5 &   27.3  & 24.4  &  34.1 	\\
BART-PT & 25.7 & 31.2 & 28.5 & 33.6 \\
\midrule
MNMT+DN (Ours) & \textbf{27.1} & \textbf{31.8} &  \textbf{29.5} & \textbf{33.7} \\
\bottomrule
\end{tabular}
\vspace{-5pt}
\caption{\label{tab:ablation}BLEU scores [\%] of the ablation study on Europarl dataset. ``+DN'' means that the experiment setting includes denoising autoencoder task. }
\end{table}

\section{Conclusion}
In this paper, we proposed to introduce denoising autoencoder objective into conventional translation objective to improve the zero-shot performance of multilingual NMT system. We analyze the motivation and effectiveness of  proposed method from the perspective of latent variables. The experimental results demonstrate that our proposed method can significantly resolve spurious correlation issue in multilingual NMT and achieves state-of-the-art performance on zero-shot translation.
In the future, it is interesting to explore the combination of our method and other language model pre-training methods~\cite{Song2019MASSMS,Liu2020MultilingualDP}. 

\section*{Acknowledgment}
We would like to thank the anonymous reviewers for the helpful comments. This work is supported by Alibaba Innovative Research Program. We appreciate Junliang Guo and Xin Zheng for the fruitful discussions. This work is done during the first author's internship at Alibaba DAMO Academy. 


\bibliography{custom}
\bibliographystyle{acl_natbib}

\appendix

\section{Appendix}
\label{sec:appendix}

\begin{table*}[t]\small
\centering
\begin{tabular}{lp{0.4cm}<{\centering}p{0.4cm}<{\centering}p{0.4cm}<{\centering}p{0.4cm}<{\centering}p{0.4cm}<{\centering}p{0.4cm}<{\centering}p{0.6cm}<{\centering}p{0.4cm}<{\centering}p{0.4cm}<{\centering}p{0.4cm}<{\centering}p{0.4cm}<{\centering}p{0.4cm}<{\centering}p{0.4cm}<{\centering}p{0.9cm}<{\centering}}
\toprule
MultiUN & \multicolumn{14}{c}{Ar, Zh, Ru $\leftrightarrow$ En} \\
\midrule
\multirow{2}*{Model} & \multicolumn{2}{c}{Ar-Ru} & \multicolumn{2}{c}{Ar-Zh} &  \multicolumn{2}{c}{Ru-Zh} & Zero &\multicolumn{2}{c}{En-Ar} & \multicolumn{2}{c}{En-Zh} &  \multicolumn{2}{c}{En-Ru} & Parallel \\
~ & $\leftarrow$ & $\rightarrow$ & $\leftarrow$ & $\rightarrow$ & $\leftarrow$ & $\rightarrow$ & Avg. & $\leftarrow$ & $\rightarrow$ & $\leftarrow$ & $\rightarrow$ & $\leftarrow$ & $\rightarrow$ & Avg. \\
\midrule
PIV-M & \textbf{29.9} & \textbf{36.8} & \textbf{29.2} & \textbf{51.5} & \textbf{34.3} & \textbf{50.1} & \textbf{38.6} & 54.7 & 37.8 & 50.7 & 58.3 & 50.7 & 42.8 & 49.2 \\
\midrule
MNMT & 17.9 & 	13.4 & 	16.1 & 	29.5 &	12.1 &	30.3 &	19.9 & 54.7 & 37.8 & 50.7 & 58.3 & 50.7 & 42.8 & 49.2 \\
LM+MNMT & 22.0 &  29.3 &	20.3 &	42.7 &	24.3 &	42.1 &	30.1 & 54.4 & 37.3 & 50.7 & 57.7 & 50.7 & 42.8 & 48.9 \\
MNMT-RS & 20.8 & 26.1 & 20.3 & 37.9 & 24.2 & 37.4 & 27.8 & 55.6 & 38.3 & 51.6 & 58.7 & 51.6 & 43.4 & 49.9  \\
BART-PT & 22.9 & 30.2 & 22.3 & 44.1 & 27.8 & 43.1 & 31.7 & 53.8 & 37.3 & 49.8 & 57.3 & 50.0 & 42.0 & 48.4\\
\midrule
MNMT+DN (Ours) & 24.6 & 33.0 & 24.6 & 47.2 & 30.0 & 46.1 &	34.3  & \textbf{56.1} & \textbf{38.0} & \textbf{52.1} & \textbf{58.6} & \textbf{52.0} & \textbf{43.9} & \textbf{50.1}\\
\bottomrule
\end{tabular}
\vspace{-5pt}
\caption{\label{tab:multiunfull}Overall BLEU scores [\%] on six zero-shot directions and six supervised directions of MultiUN dataset. ``Zero Avg.'' and ``Parallel Avg.'' refer to average BLEU score of six zero-shot directions and six supervised directions, respectively. }
\end{table*}

\begin{table*}[t]\small
\centering
\begin{tabular}{lcccccccc}
\toprule
Europarl & \multicolumn{8}{c}{De, Fr $\leftrightarrow$ En} \\
\midrule
\multirow{2}*{Model} & \multicolumn{2}{c}{De-Fr}  & Zero & \multicolumn{2}{c}{En-De} & \multicolumn{2}{c}{En-Fr} & Parallel  \\
~ & $\leftarrow$ & $\rightarrow$  & Avg. & $\leftarrow$ & $\rightarrow$  & $\leftarrow$ & $\rightarrow$ & Avg. \\
\midrule
PIV-M & 26.5 & 31.7 & 29.1 & \textbf{34.3} & \textbf{27.8} & \textbf{37.2} & \textbf{37.0} & \textbf{34.1} \\ 
\midrule
MNMT & 21.5 &   27.3  & 24.4 & \textbf{34.3} & \textbf{27.8} & \textbf{37.2} & \textbf{37.0} & \textbf{34.1}	\\
LM+MNMT & 25.5 & 31.1 &  28.3 & 33.8 & 27.2 & 36.8 & 36.4 & 33.6  \\
MNMT-RC & 25.1 & 30.8 &  28.0 & 33.5 & 27.5 & 36.6 & 36.5 & 33.5 \\
BART-PT & 25.7 & 31.2 & 28.5 & 33.6 & 27.4 & 36.8 & 36.6 & 33.6\\
\midrule
MNMT+DN (Ours) & \textbf{27.1} & \textbf{31.8} &  \textbf{29.5} & 33.8 & 27.5 & 36.8 & 36.7 & 33.7 \\
\bottomrule
\end{tabular}
\vspace{-5pt}
\caption{\label{tab:europarlfull}Overall BLEU scores [\%] on two zero-shot directions and four supervised directions of Europarl dataset. 
``Zero Avg.'' and ``Parallel Avg.'' refer to average BLEU score of two zero-shot directions and four supervised directions, respectively.}
\end{table*}

\subsection{Full Results on MultiUN and Europarl}
\label{full_result}
We report the performance of zero-shot and supervised translations on MultiUN and Europarl in Table~\ref{tab:multiunfull} and \ref{tab:europarlfull}.
We also include the pivoting version of MNMT: PIV-M.
Our proposed method still lags behind the pivoting method by an average BLEU score of 4.3 on MultiUN dataset, while achieving slightly better performance on Europarl dataset. 
Besides, our method outperforms BART pre-training by an average BLEU score of 2.6/1.0 on MultiUN and Europarl, respectively.

\subsection{Derivations for Equations}
\label{appendix_2}
The detailed derivations for latent distribution $P^{*}(Y|X;D^*,P(h|\overline{h}))$ are shown in Equation~\ref{equ:elbo}, while the derivations for the probability gap $\Delta$ in Equation~\ref{equ:delta} are as follows:
\begin{align*}
&\Delta:= P(Y|X;D^{*}) - P^{*}(Y|X;D^{*},P(h|\overline{h})) \\
& = \sum_{(x,y) \in D^{*}}{\sum_{h} P(h|\overline{h}) \log \frac{P(h|\overline{h}) P(y|x)}{P(y|h) P(h|x)}}  \\
& = \sum_{(x,y) \in D^{*}}{\sum_{h} P(h|\overline{h}) \log \frac{P(h|\overline{h}) P(y|x) P(h|y) } { P(y|h) P(h|x) P(h|y)}}  \\
& \approx \sum_{(x,y) \in D^{*}}{\sum_{h} P(h|\overline{h}) \log \frac{P(h|\overline{h}) P(y|x) P(h|y)} { P(y|h,x) P(h|x) P(h|y)}}  \\
& = \sum_{(x,y) \in D^{*}}{\sum_{h} P(h|\overline{h}) \log \frac{P(h|\overline{h}) P(y|x) P(h|y)} { P(y,h|x) P(h|y)}}  \\
& = \sum_{(x,y) \in D^{*}}{\sum_{h} P(h|\overline{h}) \log \frac{P(h|\overline{h}) P(y|x) P(h|y)} { P(h|x,y) P(y|x) P(h|y)}}  \\
& \approx \sum_{(x,y) \in D^{*}}{\sum_{h} P(h|\overline{h}) \log \frac{P(h|\overline{h}) P(y|x) P(h|y)} { P(h|y) P(y|x) P(h|y)}}  \\
& =\sum_{(x,y) \in D^{*}}{\sum_{h} P(h|\overline{h}) \log \frac{P(h|\overline{h})} {P(h|y)}}  \\
& = \sum_{(x,y) \in D^{*}} \mathbb{KL}(P(h|\overline{h})||P(h|y)),  
\end{align*}
where we use two approximations here that are $P(y|h,x)\approx P(y|h)$ and $P(h|x,y)\approx P(h|y)$.

\end{document}